# 蝴蝶种类自动识别研究

谢娟英 [1] 侯琦 [1] 史颖欢 [2] 吕鹏 [3] 景丽萍 [4] 庄福振 [5] 张军平 [6] 谭晓阳 [7] 许升全 [8]

[1]（陕西师范大学 计算机科学学院，陕西 西安 710119）
[2]（南京大学 计算机科学与技术系，江苏 南京 210023）
[3]（山东财经大学 计算机科学与技术学院，山东 济南 250014）
[4]（北京交通大学 计算机科学与技术学院，北京 100044）
[5]（中国科学院计算技术研究所，北京 100190）
[6]（复旦大学 计算机科学与技术学院，上海 200433）
[7]（南京航空航天大学 计算机科学与技术学院，南京 210016）
[8]（陕西师范大学 生命科学学院，陕西 西安 710119）
（通讯作者：许升全， email：xiejuany@snnu.edu.cn）

# The Automatic Identification of Butterfly Species

XIE Juan-Ying[1], HOU Qi[1], SHI Ying-Huan [2], LV Peng [3], JING Li-Ping [4], ZHUANG Fu-Zhen[5], ZHANG Jun-Ping [6], TAN Xiao-Yang [7], XU Sheng-Quan[8]

[1]（*School of Computer Science, Shaanxi Normal University, Xi'an 710062*）
[2]（*Department of Computer Science & Technology, Nanjing University, Nanjing 210023*）
[3]（*School of Computer Science & Technology, Shandong University of Finance and Economics, Ji'nan 250014*）
[4]（*School of Computer & information Technology, Beijing Jiaotong University, Beijing 100044*）
[5]（*Institute of computing technology, Chinese Academy of Sciences，Beijing 100190*）
[6]（*School of Computer Science, Fudan University，Shanghai 200433*）
[7]（*College of Computer Science and Technology, Nanjing University of Aeronautics and Astronautics, Nanjing 210016*）
[8]（*College of Life Sciences, Shaanxi Normal University, Xi'an 710062*）

**Abstract** The available butterfly data sets comprise a few limited species, and the images in the data sets are always standard patterns without the images of butterflies in their living environment. To overcome the aforementioned limitations in the butterfly data sets, we build a butterfly data set composed of all species of butterflies in China with 4270 standard pattern images of 1176 butterfly species, and 1425 images from living environment of 111 species. We propose to use the deep learning technique Faster-Rcnn to train an automatic butterfly identification system including butterfly position detection and species recognition. We delete those species with only one living environment image from data set, then partition the rest images from living environment into two subsets, one used as test subset, the other as training subset respectively combined with all standard pattern butterfly images or the standard pattern butterfly images with the same species of the images from living environment. In order to construct the training subset for Faster-Rcnn, nine methods were adopted to amplifying the images in the training subset including the turning of up and down, and left and right, rotation with different angles, adding noises, blurring, and contrast ratio adjusting etc. Three prediction models were trained. The mAP (Mean Average prediction) criterion was used to evaluate the performance of the prediction model. The experimental results demonstrate that our Faster-Rcnn based butterfly automatic identification system performed well, and its worst mAP is up to 60％, and can simultaneously detect the positions of more than one butterflies in one images from living environment and recognize the species of those butterflies as well.

**摘要** 针对现有蝴蝶识别研究中所用数据集蝴蝶种类偏少，且只含有蝴蝶标本照片，不含生态环境中蝴蝶照片的问题，发布了一个同时包含标本照片和生态照片的蝴蝶数据集，其中的标本照片包含中国全部蝴蝶种类，共

---

计 4270 张照片，1176 种；蝴蝶生态环境下的照片 1425 张，111 种。提出基于深度学习技术 Faster-Rcnn 的蝴蝶种类自动识别技术，包括生态照片中蝴蝶位置的自动检测和物种鉴定。实验去除只含有单张生态照片的蝴蝶种类后，对剩余的蝴蝶生态照片进行 5-5 划分，构造 2 种不同训练数据集：一半生态照片+全部模式照片，一半生态照片+对应种类模式照片；训练 3 种不同网络结构的蝴蝶自动识别系统，以 mAP( Mean Average Precision) 为评价指标，采用上下、左右翻转、不同角度旋转、加噪、不同程度模糊、对比度升降等 9 种方式扩充训练集。实验结果表明，基于 Faster-Rcnn 深度学习框架的蝴蝶自动识别系统对生态环境中的蝴蝶照片能实现自动检测和物种识别，模型的 mAP 最低值接近 60%，并能同时检测出生态照片中的多只蝴蝶和完成物种识别。

关键词　蝴蝶；自动识别；目标检测；深度学习；分类
中图法分类号　　TP181

蝴蝶是节肢动物门、昆虫纲、鳞翅目、锤角亚目昆虫的统称，全世界大约有 18,000 多种[1]，中国的蝴蝶种类约有 1,200 种。蝴蝶大多白天活动，其翅和身体一般都有各色鳞片形成的色斑和花纹。这些色彩和斑纹往往和蝴蝶的生活环境一致，可以避免被鸟类等蝴蝶天敌发现，是蝴蝶的保护色。蝴蝶自古以来就是重要的文化昆虫，不仅美化人类生活，梁祝化蝶等爱情故事千百年来也被人们传颂。同时多数蝴蝶对其生存的自然环境要求颇高，因此是重要的环境指示昆虫。但人工鉴别蝴蝶种类不仅需要长期的经验积累而且费时费力，严重影响了人类对蝴蝶的认识。为了提高蝴蝶物种识别效率，方便众多昆虫爱好者及大众了解和认识蝴蝶，本文将建立包含蝴蝶生态照片和标本模式照片的中国蝴蝶数据全集；提出基于深度学习模型 Faster-Rcnn 的生态照片蝴蝶自动检测和物种自动鉴定方法，构建生态照片蝴蝶种类自动识别系统。

## 1 蝴蝶自动鉴定研究背景

近年来，蝴蝶自动识别受到了越来越多研究者关注，出现了大量蝴蝶识别应用研究。在基于内容检索的蝴蝶所属科识别[2]中根据蝴蝶形状，颜色，纹理分别提取图像特征，然后根据相似度返回相应分类结果。基于极限学习机的蝴蝶种类自动识别[3]使用 local binary patterns [4] (LBP) 和 grey-level co-occurrence matrix [5] (GLCM) 提取蝴蝶图片纹理特征，然后使用极限学习机进行分类。基于单隐层神经网络的蝴蝶识别[6]使用 BLS(branch length similarity)提取蝴蝶形状特征，然后使用单隐层神经网络进行分类。Seung-Ho Kang 提出了借助多视角观察蝴蝶图片来扩充训练集，以训练神经网络的蝴蝶识别方法[7]。基于神经网络的物种自动识别[8]，从图片的几何结构、形态学、纹理特征三方面对鱼、蝴蝶、植物三类不同生物设计了 15 个特有特征，然后使用神经网络进行训练和物种分类。现有蝴蝶物种识别研究存在如下问题：第一，使用的蝴蝶数据集均只包含较少的蝴蝶种类；第二，使用的蝴蝶照片均是标准蝴蝶照片，即蝴蝶标本的模式照片，没有涉及蝴蝶在自然生态环境中的生态照片。鉴于此，本研究将提供一个较完整的蝴蝶照片数据集，包含中国现有的所有蝴蝶标本标准照片，以及尽可能多的蝴蝶在其生态环境中的照片，并使该数据集可随时间积累不断完善。

深度学习作为机器学习领域的一项技术突破，在图像处理领域表现出非常好的潜能。为此，本研究提出采用深度学习技术，依据在自然界拍摄的蝴蝶生态照片对蝴蝶种类进行自动识别研究，克服现有蝴蝶自动识别研究只依据蝴蝶标本标准照片进行蝴蝶种类识别的缺憾，提供具有挑战性的真正意义上的蝴蝶自动识别。

本研究通过野外拍摄、蝴蝶爱好者提供、扫描中国蝶类志[9]中的蝴蝶照片等方式收集了包含中国所有蝴蝶种类的蝴蝶照片数据集。为了实现根据蝴蝶生态照片自动识别蝴蝶种类，需要完成生态环境照片中蝴蝶的定位问题，以及在此基础上的蝴蝶自动分类识别问题。为此，本文将使用深度学习技术，同时完成蝴蝶生态环境照片中的蝴蝶定位和分类识别。

## 2 数据与方法

### 2.1 蝴蝶图像数据集

本文构建的蝴蝶数据集的蝴蝶照片不仅包括标准的蝴蝶标本照片，还包括自然生态环境中的蝴蝶照片。数据集中的蝴蝶标本标准照片，称为模式照片，主要来源于中国蝴蝶蝶类志[9]；蝴蝶在其生态环境中的照片，简称生态照，来源于野外实地拍摄和蝴蝶爱好者捐赠。

数据集共包含 5,695 张蝴蝶照片，其中的生态照片如图 1 所示，共有 1,425 张，111 种，均为在野外使用高清单反相机拍摄所得。其中 17 种蝴蝶的生态照片只有一张，即有 17 种蝴蝶的生态照片只有一个样本。蝴蝶生态照片的特点是每张照片包含的蝴蝶数量、种类不确定，而且蝴蝶由于自身的拟态躲避天敌

的缘故，总是偏向于处在有利于隐藏自身的环境中，使得蝴蝶和周围环境较难辨别。生态照片的数据统计见图 2 所示，多数生态照的蝴蝶其类别包含的样本数量在 20 个以内，每个蝴蝶种类至少包含一个样本，最多的包含有 121 个样本，呈现出典型的长尾分布。

蝴蝶标本的标准照片，即模式照，如图 3 所示，通过扫描中国蝶类志[9]的蝴蝶图片得来，共有 4,270 张，1176 种，包含生态照的所有种类。模式照的特点是一张图片确定地含有一只蝴蝶。模式照的数据分布如图 4 所示，多数模式照蝴蝶类别包含的样本数量在 4 个左右，每个蝴蝶种类至少包含一个样本，最多含有 14 个样本。对比图 4 和图 2 可见，模式照数据集包含的蝴蝶种类多，每类样本数量相差不大，但依然呈现出典型的长尾分布。我们将模式照的蝴蝶种类按照科、亚科、属、种、亚种，以及蝴蝶的雌、雄、背、腹进行分类。由于部分亚种缺失，部分蝴蝶种类雌、雄、背或腹不全。因此，为了统一，实验中将蝴蝶数据集所有模式照的蝴蝶和生态照的蝴蝶均精确到种。该数据集将在"**2018 年第三届中国数据挖掘竞赛——国际首次蝴蝶识别大赛**"中首次对外公开。

数据集中的模式照包含中国现有全部蝴蝶种类，构成蝴蝶种类全集，生态照的蝴蝶种类是模式照蝴蝶种类的子集。整理后的数据集共有 1,176 种蝴蝶，其中 48 种蝴蝶的照片只有一张照片，即共有 48 种蝴蝶只有一个样本。我们的目标是自动识别蝴蝶生态照片中的蝴蝶种类。因此，以生态照为准，去掉只有单张生态照的蝴蝶类别图片，最终的实验数据中，生态照有 1408 张图片，94 种蝴蝶。按照训练集、测试集各 50%的比例进行划分，训练集向上取整，得到包括 721 张蝴蝶生态照片的训练集，包括 687 张蝴蝶生态照片的测试集。图 5 表示训练集和测试集的蝴蝶生态照片数量统计分布，可以看出训练集和测试集的分布大致相同，以确保现有数据环境下得到尽可能优的结果。

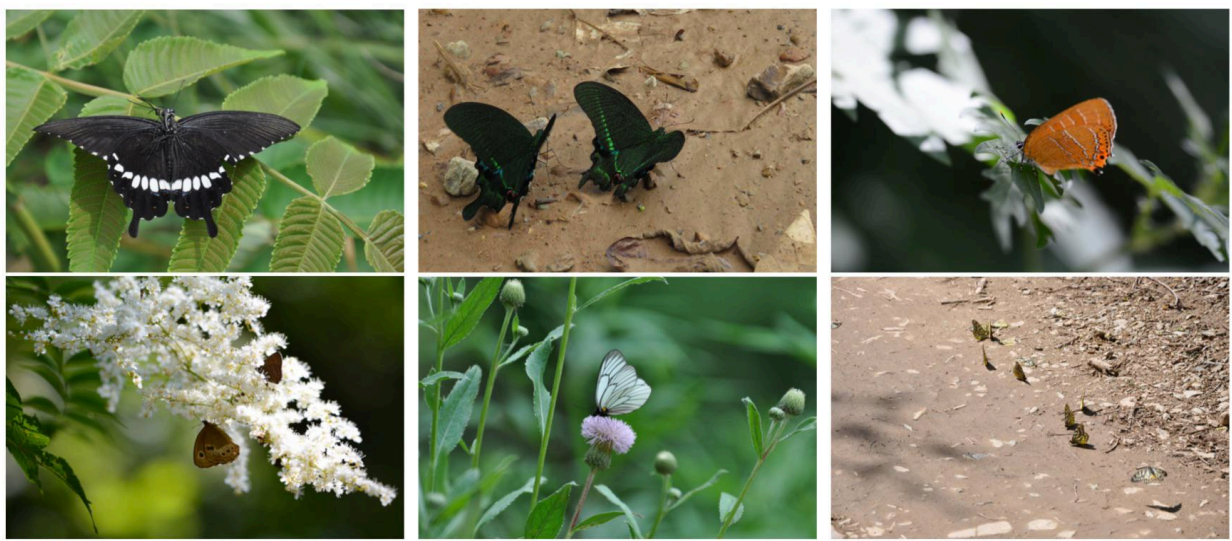

Fig. 1 Samples of natural butterfly images
图 1 生态照的部分样本

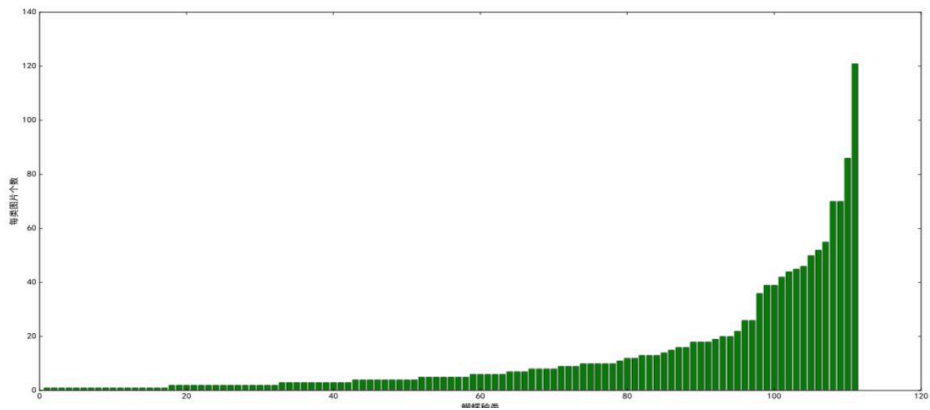

Fig. 2 Data distribution of natural butterfly images
图 2 生态照数据分布图

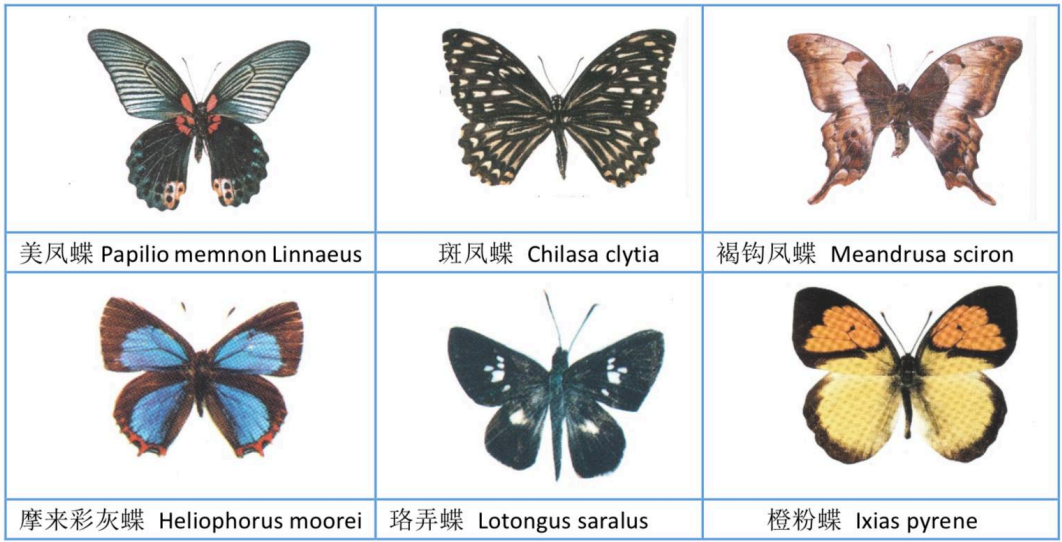

Fig. 3 Samples of butterfly pattern images
图 3 模式照的部分样本

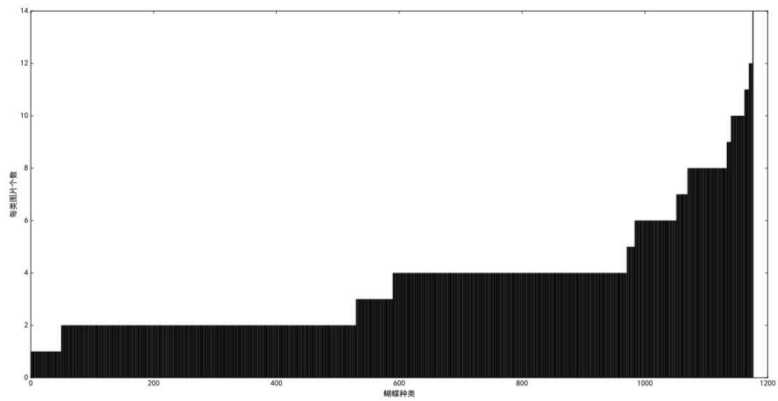

Fig. 4 Data distribution of standard pattern images of butterflies
图 4 蝴蝶模式照数据分布图

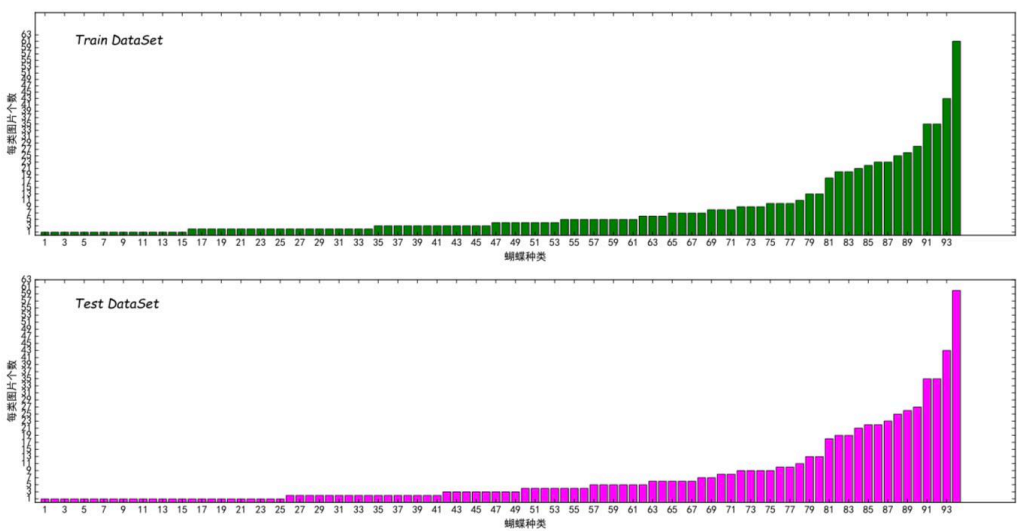

Fig. 5 Data distribution of natural butterfly images on train and test dataset
图 5 蝴蝶生态照训练集和测试集数据分布图

## 2.2 数据集划分

我们对所有生态照的蝴蝶位置进行人工标注，模式照的蝴蝶位置默认为全图大小。由于深度学习需要大量训练数据，因此对训练数据集图片采用翻转、旋

转、加噪、模糊、对比度升降等 9 种方式进行变换，以扩充训练数据集的蝴蝶图片数量。

训练集图片同时包含生态照和模式照。所有生态照在去掉只有一个样本的蝴蝶种类后，划分训练集和测试集。我们的目标是对生态照中的蝴蝶，同时进行定位和分类，因此，测试集只包含蝴蝶生态照片。我们希望借助模式照的一些信息预测生态照的蝴蝶类别，因此，将模式照图片也加入训练集。模式照图片加入训练集的方式分为两种，第一种是将所有模式照都加入训练集，这样做是考虑到蝴蝶全集可以更好的有助于提取蝴蝶间的共有特征信息，另外还可以提高数据的扩展性，提供包含所有蝴蝶种类的预训练版本，如果有新数据增加，可以在这个版本上继续进行增强训练；第二种方式是只加入与生态照种类对应的模式照图片，这是更为通用的方法，只对要分类的蝴蝶图片进行训练，减小模型的复杂度，提高分类精度。如 2.1 节所述，生态照在去掉只有一个样本的蝴蝶种类后，包含 94 种蝴蝶，1,408 张蝴蝶生态图片。按照训练集和测试集各 50% 的比例划分生态照，测试集包括 687 张生态照，其余 721 张生态照片加入训练集。按照第一种构造训练集的方法，将所有模式照加入训练集，再对训练集所有图片进行扩充，最终得到训练集的蝴蝶图片为 49,910 张，我们称之为 Data_1。按照第二种生成训练集的方式，训练集只加入生态照蝴蝶种类对应的蝴蝶模式照，得到蝴蝶训练数据集的图片 13,060 张，命名为 Data_2。至此，我们将蝴蝶自动识别问题转化为一个 94 类的多类目标自动检测和识别问题。与普通的多类目标检测和识别问题相比，我们的蝴蝶自动检测与识别问题的难点不仅在于类别多，更重要的是我们要进行的是相同大类（蝴蝶）下的小类识别，或称为细粒度分类，因此，本文的蝴蝶自动识别研究更具有挑战性。

**2.3 蝴蝶位置检测与种类识别方法**

本文采用 Girshick R 等人[10]提出的目标检测方法，同时实现目标的定位和分类。Faster-Rcnn[10]结构如图 6 所示，它是对 Rcnn[11]和 Fast-Rcnn[12]的改进，相比于开山之作的 Rcnn，Fast-Rcnn 提出新的 ROI(Region Of Interest)层来规避冗余的特征提取方式，只对整张图像全区域进行一次特征提取，同时加入了多任务学习，使得网络训练过程中同时学习物体分类和窗口位置回归。Faster-Rcnn 在 Fast-Rcnn 基础上加入了区域生成网络层 RPN(Region Proposal Networks)，来代替耗时的候选区域生成方式，真正实现端到端的目标检测任务。

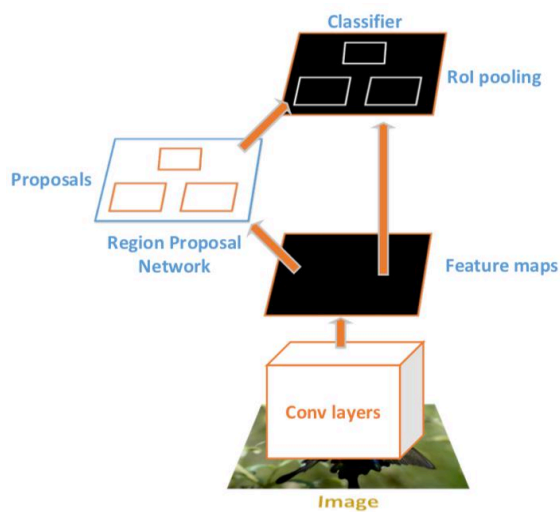

Fig. 6 Faster-RCNN network topology
图 6 Faster RCNN 的网络结构

Faster-Rcnn 区域生成网络 RPN 采用多任务学习，即将生态照中蝴蝶的位置定位和蝴蝶种类鉴定两个任务同时进行。式(1)表示 RPN 网络多任务损失函数，即分类损失和回归损失之和，其中 $i$ 为一个参考窗口(anchor)在一个随机梯度下降的小批量分层采样(SGD mini-batch)中的下标。$p_i$ 表示预测第 $i$ 个 anchor 为某类样本的概率。$p_i^*$ 为相应样本的真实类别标签。如果这个参考窗口是正样本的，则 $p_i^*$ 为 1，否则为 0。$t_i$ 表示正样本 anchor 到预测区域(bounding box)的 4 个参数化坐标，$t_i^*$ 是这个正样本 anchor 对应的真实类别位置区域的标签(ground-truth)。 $L_{cls}$ 是一个二值分类器的分类损失，$L_{reg}$ 通过 $L_{reg}(t_i, t_i^*) = R(t_i - t_i^*)$ 计算回归损失，其中函数 R 是式(2)所示的 smooth $L_1$ 函数。$\lambda$ 为权重参数，$N_{cls}$ 和 $N_{reg}$ 用来标准化两个损失项。

$$L(\{p_i\},\{t_i\}) = \frac{1}{N_{cls}}\sum_i L_{cls}(p_i, p_i^*) + \lambda \frac{1}{N_{reg}}\sum_i p_i^* L_{reg}(t_i, t_i^*) \quad (1)$$

$$\text{smooth}_{L_1}(x) = \begin{cases} 0.5 x^2 & if\ |x| < 1 \\ |x| - 0.5 & otherwise \end{cases} \quad (2)$$

Faster-Rcnn 图片检测流程如图 7 所示，我们首先通过共享卷积层提取蝴蝶图片特征，得到图片的特征映射，输入的特征图通过滑动窗口扫描，每个滑窗位置得到 9 种可能的参考窗口(anchors)，得到区域建议的低维特征向量，再经过 RPN 网络得到区域建议和区域得分，并对区域得分采用非极大值抑制，输出 Top-N 得分的区域建议给 RoI 池化层；通过 ROI 池化层得到区域建议特征，区域建议特征通过全连接层后，输出该区域的分类得分以及区域位置。

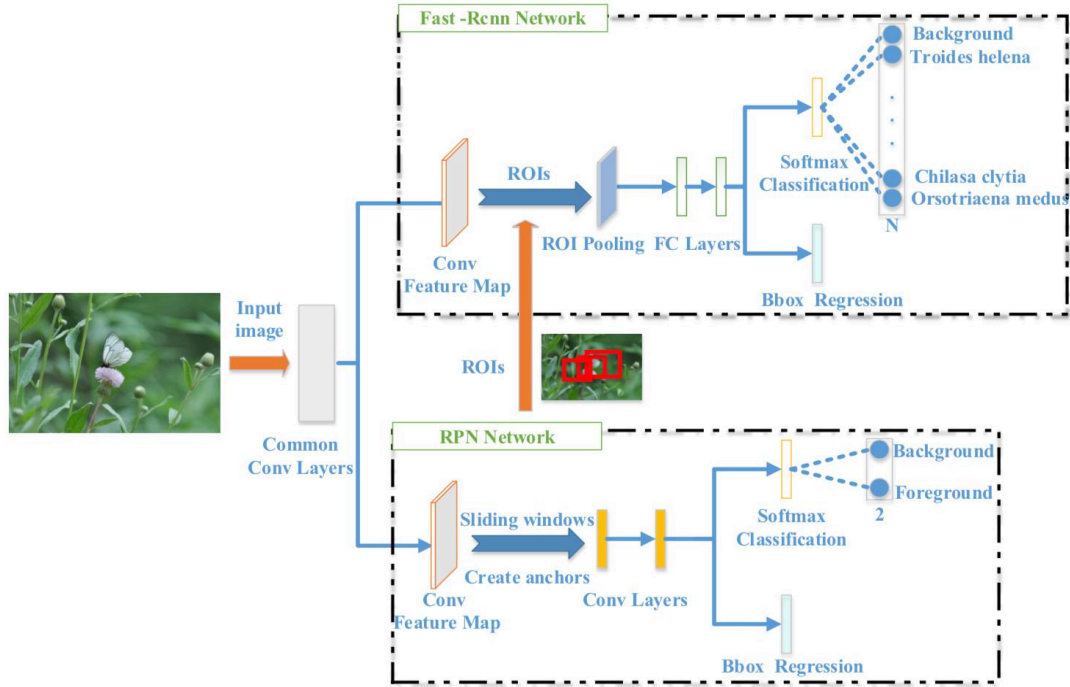

Fig. 7 Faster-Rcnn image detection procedure

图 7 Faster-Rcnn 图片检测流程

实验采用 ZF[13]，VGG_CNN_M_1024[14]，VGG16[15]三种网络结构，使用**联合训练方式**，分别训练了 3 个蝴蝶自动识别系统。实验首先在 ImageNet 数据上进行预训练，然后使用扩充的蝴蝶训练数据集进行微调，所有模型的超参数使用默认参数，初始学习率为 0.001，随后按步长调整；动量设置为 0.9，权重衰减设置为 0.0005，共迭代 10W 次。

**2.4 蝴蝶识别的评价指标**

使用 IOU (Intersection Over Union)作为区域检测的评价指标，其定义是预测产生的候选框(candidate region)与原标记框(ground-truth region)的交叠率，即它们的交集与并集的比值。最理想情况是完全重叠，即比值为 1，实验中取 IOU>0.5。我们的目标是生态照中蝴蝶种类的正确识别，因此，实验结果中没有具体列出生态照中蝴蝶位置检测结果的评价指标值。

多类图像分类评价指标 mAP (Mean Average Precision) 作为目标检测的通用指标，是评价多类分类的平均准确性，因此，本文采用 mAP 作为蝴蝶识别情况的评价指标。mAP 根据精度(Precision)和召回率(Recall)计算得到。精度的计算方法如式(3)所示，召回率的计算方法如式(4)所示，其中 TP 为预测结果中正确分类和定位的正样本个数，TN 为正确分类和定位的负样本个数，FP 为负样本被错误标记为正样本的个数，FN 是正样本被错误标记为负样本的个数。

$$precision = \frac{TP}{TP+FP} \quad (3)$$

$$recall = \frac{TP}{TP+FN} \quad (4)$$

为预测训练所得模型的性能，我们定义正样本为：真实框(ground-truth)和预选框的交并比大于 0.5，且预测的分类概率大于 0.5，也就是定位和分类同时满足条件才可认为是正样本。

根据不同的置信度可以得到若干个(Precision, Recall)点，以召回率为横坐标，精度为纵坐标画出 P-R 曲线，AP(Average Precision)是精度和召回率曲线下的面积，也就是 P-R 曲线的积分，如式 (5)所示。实际计算中，一般都用若干矩形面积来代替曲线下面积，即将召回率划分为 $n$ 块，$[0,1/n,\cdots,(n-1)/n,1]$，则 AP 可以表示为式(6)。最终得到的所有类的平均精度，如式(7)所示，其中 N 为测试数据中的蝴蝶种类数。对于具体的矩形划分，采用 PASCAL VOC Challenge 在 2010 年以后的计算方法[16]。

$$AveP = \int_0^1 p(r)dr \quad (5)$$

$$AveP = \frac{1}{n}\sum_{i=1}^{n}\max_{r\in[(i-1)/n,i/n]}p(r) \quad (6)$$

$$mAP = \frac{\sum_{n=1}^{N}AveP(n)}{N} \quad (7)$$

## 3 实验结果与分析

对两种不同的训练数据集和三种不同的网络结构，最终得到测试集蝴蝶种类识别的 mAP 结果如表 1 所示。测试集上蝴蝶位置检测和分类鉴定的实际效果如图 8 所示。由于空间有限，图 8 只列出了部分测试结果。

**Table 1 mAP Results on Test Data with Different Train Data of butterflies**

**表 1 不同蝴蝶训练集下的蝴蝶测试集的 mAP 结果**

| Network / Train Data | ZF | VGG_CNN_M_1024 | VGG16 |
| --- | --- | --- | --- |
| Data_1 | 59.8% | 64.5% | 72.8% |
| Data_2 | 73.3% | 72.6% | 76.1% |

从表 1 实验结果可以看出，无论采用哪种网络结构，基于 Faster-Rcnn 的蝴蝶自动识别系统对生态照片中的蝴蝶种类识别均显示出不错的效果，最差的 mAP 值也接近 60%，这不仅说明 Faster-Rcnn 在细粒度分类的性能，也说明提出的基于 Faster-Rcnn 的蝴蝶生态照中蝴蝶种类自动识别的可行性，同时也说明了生态照片中蝴蝶位置检测的准确性。另外，表 1 数据还揭示，Data_2 所得模型的 mAP 值在整体上优于 Data_1 所得模型的 mAP 值，这说明训练集包含 721 张蝴蝶生态照片及与这些生态照蝴蝶种类对应的蝴蝶模式照时，训练所得模型的蝴蝶位置自动检测与蝴蝶种类自动识别能力优于包含 721 张蝴蝶生态照与模式照全集所得模型的性能。只加入与生态照种类对应的模式照片的性能更优的原因在于其降低了模型的复杂度，减少了过多无用蝴蝶种类的干扰，因此其分类准确率更高。表 1 关于 3 种不同网络结构的实验数据还可见：在蝴蝶生态照自动检测与分类识别中，VGG16 网络的性能优于其他两种网络，但是该模型的复杂度高，训练时间长，故实际使用过程中可选用分类和位置检测性能较优，训练速度较快的中型网络 VGG_CNN_M_1024。

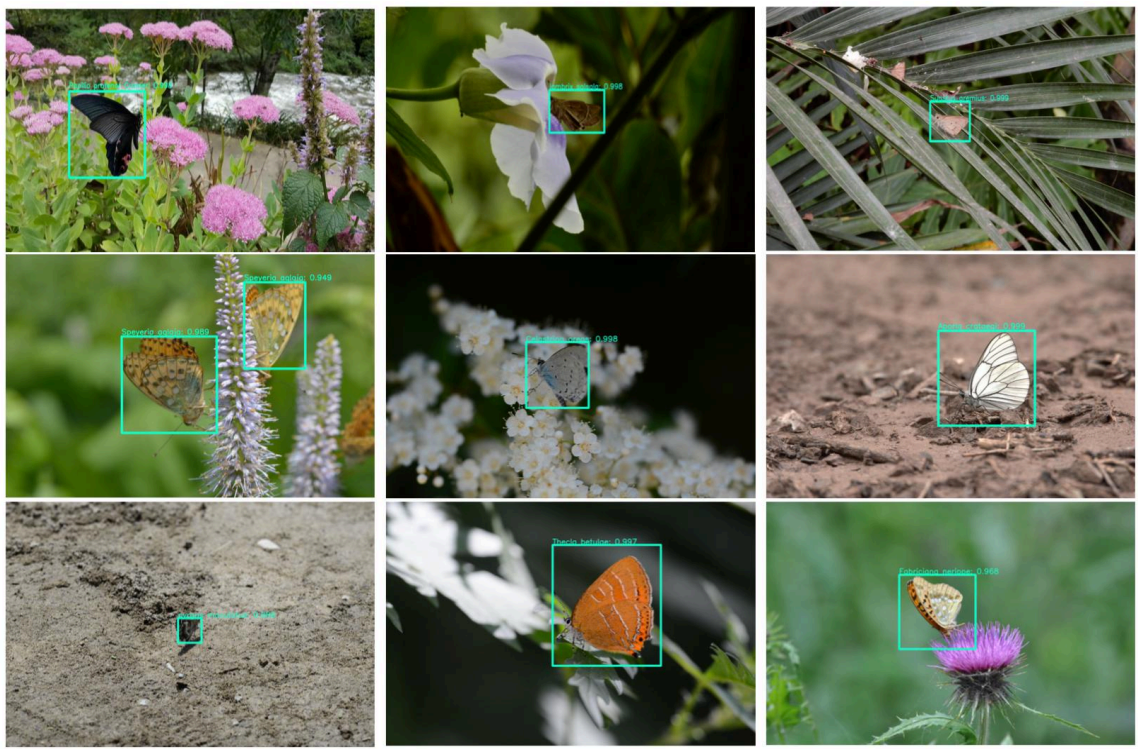

Fig. 8 The Detection Results on natural images

图 8 部分生态照的检测结果

图 8 所示部分生态照蝴蝶位置检测和种类识别结果显示：提出的基于 Faster-Rcnn 的蝴蝶自动识别模型，可以同时实现蝴蝶生态照片中蝴蝶的位置定位和种类分类识别；还可以同时定位出生态照片中的多只蝴蝶，并正确分类识别其种类；对非常拟态的蝴蝶生态照片中的蝴蝶也能进行位置检测和物种鉴定。

## 4 结论

本文发布了一个全新的蝴蝶数据集，填补了现有蝴蝶自动识别研究使用的蝴蝶数据集没有自然场景下的蝴蝶数据的空白，同时提供了一个完整的中国现有蝴蝶数据全集，供"**2018 年第三届中国数据挖掘竞赛——国际首次蝴蝶识别大赛**"使用，并可供所有对蝴蝶自动识别感兴趣的研究者们使用，也可用于测试

目标自动检测和分类的相关算法。另外，本文基于 Faster-Rcnn 深度学习模型，构建了一个对自然场景下拍摄的蝴蝶照片中的蝴蝶进行自动定位和物种自动分类鉴定的蝴蝶自动识别系统。最后，本文提出的数据集不仅可用于目标检测领域，还可以用于小样本分类，细粒度分类等方向，这也是我们未来的研究工作之一。